# A Bayesian Methodology for Estimating Uncertainty of Decisions in Safety-Critical Systems

Vitaly SCHETININ[a,1], Jonathan E. FIELDSEND[b], Derek PARTRIDGE[b], Wojtek J. KRZANOWSKI[b], Richard M. EVERSON[b], Trevor C. BAILEY[b] and Adolfo HERNANDEZ[b]

[a]*Department of Computing and Information Systems, University of Luton, LU1 3JU, UK*
[b]*School of Engineering, Computer Science and Mathematics, University of Exeter, EX4 4QF, UK*

**Abstract.** Uncertainty of decisions in safety-critical engineering applications can be estimated on the basis of the Bayesian Markov Chain Monte Carlo (MCMC) technique of averaging over decision models. The use of decision tree (DT) models assists experts to interpret causal relations and find factors of the uncertainty. Bayesian averaging also allows experts to estimate the uncertainty accurately when *a priori* information on the favored structure of DTs is available. Then an expert can select a single DT model, typically the Maximum a Posteriori model, for interpretation purposes. Unfortunately, *a priori* information on favored structure of DTs is not always available. For this reason, we suggest a new prior on DTs for the Bayesian MCMC technique. We also suggest a new procedure of selecting a single DT and describe an application scenario. In our experiments on real data our technique outperforms the existing Bayesian techniques in predictive accuracy of the selected single DTs.

**Keywords.** Uncertainty, decision tree, Bayesian averaging, MCMC.

## Introduction

The assessment of uncertainty of decisions is of crucial importance for many safety-critical engineering applications [1], e.g., in air-traffic control [2]. For such applications Bayesian model averaging provides reliable estimates of the uncertainty [3, 4, 5]. In theory, uncertainty of decisions can be accurately estimated using Markov Chain Monte Carlo (MCMC) techniques to average over the ensemble of diverse decision models. The use of decision trees (DT) for Bayesian model averaging is attractive for experts who want to interpret causal relations and find factors to account for the uncertainty [3, 4, 5].

Bayesian averaging over DT models allows the uncertainty of decisions to be estimated accurately when *a priori* information on favored structure of DTs is available as described in [6]. Then for interpretation purposes, an expert can select a single DT model which provides the Maximum a Posteriori (MAP) performance [7]. Unfortunately, in most practical cases, *a priori* information on the favored structure of DTs is not available. For this reason, we suggest a new prior on DT models within a sweeping strategy that we described in [8].

---
[1]Corresponding Author: Vitaly Schetinin, Department of Computing and Information Systems, University of Luton, Luton, LU1 3JU, The UK; E-mail: vitaly.schetinin@luton.ac.uk.

We also suggest a new procedure for selecting a single DT, described in Section 3. This procedure is based on the estimates obtained within the Uncertainty Envelope technique that we described in [9]. An application scenario, which can be implemented within the proposed Bayesian technique, is described in Section 5.

In this Chapter we aim to compare the predictive accuracy of decisions obtained with the suggested Bayesian DT technique and the standard Bayesian DT techniques. The comparison is run on air-traffic control data made available by the National Air Traffic Services (NATS) in the UK. In our experiments, the suggested technique outperforms the existing Bayesian techniques in terms of predictive accuracy.

## 1. Bayesian Averaging over Decision Tree Models

In general, a DT is a hierarchical system consisting of splitting and terminal nodes. DTs are binary if the splitting nodes ask a specific question and then divide the data points into two disjoint subsets [3]. The terminal node assigns all data points falling in that node to the class whose points are prevalent. Within a Bayesian framework, the class posterior distribution is calculated for each terminal node, which makes the Bayesian integration computationally expensive [4].

To make the Bayesian averaging DTs a feasible approach, Denison *et al.* [5] have suggested the use of the MCMC technique, taking a stochastic sample from the posterior distribution. During sampling, the parameters $\theta$ of candidate-models are drawn from the given proposal distributions. The candidate is accepted or rejected accordingly to Bayes rule calculated on the given data **D**. Thus, for the *m*-dimensional input vector **x**, data **D** and parameters $\theta$, the class posterior distribution $p(y|\mathbf{x},\mathbf{D})$ is

$$p(y|\mathbf{x},\mathbf{D}) = \int p(y|\mathbf{x},\theta,\mathbf{D})p(\theta|\mathbf{D})d\theta \approx \frac{1}{N}\sum_{i=1}^{N} p(y|\mathbf{x},\theta^{(i)},\mathbf{D}),$$

where $p(\theta|\mathbf{D})$ is the posterior distribution of parameters $\theta$ conditioned on data **D**, and *N* is the number of samples taken from the posterior distribution.

Sampling across DT models of variable dimensionality, the above technique exploits a Reversible Jump (RJ) extension suggested by Green [10]. When *priori* information is not distorted and the number of samples is reasonably large, the RJ MCMC technique, making birth, death, change-question, and change-rule moves, explores the posterior distribution and as a result provides accurate estimates of the posterior.

To grow large DTs from real-world data, Denison *et al.* [5] and Chipman *et al.* [6] suggested exploring the posterior probability by using the following types of moves:

*Birth*. Randomly split the data points falling in one of the terminal nodes by a new splitting node with the variable and rule drawn from the corresponding priors.

*Death*. Randomly pick a splitting node with two terminal nodes and assign it to be a single terminal with the united data points.

***Change-split***. Randomly pick a splitting node and assign it a new splitting variable and rule drawn from the corresponding priors.

***Change-rule***. Randomly pick a splitting node and assign it a new rule drawn from a given prior.

The first two moves, *birth* and *death*, are reversible and change the dimensionality of $\theta$ as described in [10]. The remaining moves provide jumps within the current dimensionality of $\theta$. Note that the *change-split* move is included to make "large" jumps which potentially increase the chance of sampling from a maximal posterior whilst the *change-rule* move does "local" jumps.

For the birth moves, the proposal ratio $R$ is written

$$R = \frac{q(\theta | \theta') p(\theta')}{q(\theta' | \theta) p(\theta)},$$

where $q(\theta | \theta')$ and $q(\theta' | \theta)$ are the proposed distributions, $\theta'$ and $\theta$ are $(k + 1)$ and $k$-dimensional vectors of DT parameters, respectively, and $p(\theta)$ and $p(\theta')$ are the probabilities of the DT with parameters $\theta$ and $\theta'$:

$$p(\theta) = \left\{ \prod_{i=1}^{k-1} \frac{1}{N(s_i^{var})} \frac{1}{m} \right\} \frac{k}{S_k} \frac{1}{K},$$

where $N(s_i^{var})$ is the number of possible values of $s_i^{var}$ which can be assigned as a new splitting rule, $S_k$ is the number of ways of constructing a DT with $k$ terminal nodes, and $K$ is the maximal number of terminal nodes, $K = n - 1$.

The proposal distributions are as follows

$$q(\theta | \theta') = \frac{d_{k+1}}{D_{Q'}},$$

where $D_{Q1} = D_Q + 1$ is the number of splitting nodes whose branches are both terminal nodes.

Then the proposal ratio for a *birth* is given by

$$R = \frac{d_{k+1}}{b_k} \frac{k}{D_{Q1}} \frac{S_k}{S_{k+1}}.$$

The number $D_{Q1}$ is dependent on the DT structure and it is clear that $D_{Q1} < k \ \forall \ k = 1, \ldots, K$. Analyzing the above equation, we can also assume $d_{k+1} = b_k$. Then letting the DTs grow, i.e., $k \to K$, and considering $S_{k+1} > S_k$, we can see that the value of $R \to c$, where $c$ is a constant lying between 0 and 1.

Alternatively, for the death moves the proposal ratio is written as

$$R = \frac{b_k}{d_{k-1}} \frac{D_Q}{(k-1)} \frac{S_k}{S_{k-1}}.$$

However, in practice the lack of *a priori* information brings bias to the posterior estimates, and as a result the evaluation of classification uncertainty may be incorrect [11].

Within the RJ MCMC technique, the prior on the number of splitting nodes should be given properly. Otherwise, most samples may be taken from the posterior calculated for DTs that are located far away from a region containing the desired DT models. Likewise, when the prior on the number of splits is assigned as uniform, the minimal number of data points, $p_{min}$, allowed to be at nodes may be set inappropriately small. In this case, the DTs will grow excessively and most of the samples will be taken from the posterior distribution calculated for over-fitted DTs. As a result, the use of inappropriately assigned priors leads to poor results [5, 6].

For the special cases when there is knowledge of the favored DT structure, Chipman *et al.* [6] suggested the prior probability, with which a terminal node should be split further. This probability is dependent on how many splits have been made above it. For the given constants $\gamma > 0$ and $\delta \geq 0$, the probability $P_s$ of splitting the *i*th node is

$$P_s(i) = \gamma(1 + d_i)^{-\sigma},$$

where $d_i$ is the number of splits made above node *i*. Here the additional parameters $\gamma$ and $\delta$ serving as hyperpriors should be given properly.

## 2. A Sweeping Strategy

Clearly, the lack of *a priori* knowledge on the favored DT structure, which often happens in practice, increases the uncertainty in results of the Bayesian averaged DTs. To decrease the uncertainty of decisions, a new Bayesian strategy of sampling DT models has been suggested [8]. The main idea behind this strategy is to assign the *a priori* probability of further splitting DT nodes dependent on the range of values within which the number of data points will be not less than $p_{min}$. This prior is explicit because at the current partition the range of such values is unknown.

Within the above prior, the new splitting value $q_j'$ for variable *j* is drawn from a uniform distribution:

$$q_j' \sim U(x_{\min}^{1,j}, x_{\max}^{1,j}),$$

and from a Gaussian with a given variance $\delta_j$:

$$q_j' \sim N(q_j, \delta_j),$$

for the birth and change moves, respectively.

Because of the hierarchical structure, new moves applied to the first partition levels can heavily change the shape of the DT and, as a result, at its bottom partitions the terminal nodes can contain fewer data points than $p_{min}$. However, if the DT contains one such node, we can sweep it and then make the death move. Less likely, after birth or change moves the DT will contain more than one node containing fewer than $p_{min}$ data points. In such cases we have to resample the DT.

**3. Selection of a Single DT**

In this Section we describe our method of interpreting Bayesian DT ensembles. This method is based on the estimates of confidence in the outcomes of the DT ensemble which can be quantitatively estimated on the training data within the Uncertainty Envelope technique.

*3.1. Selection Techniques*

There are two approaches to interpreting DT ensembles. The first approach is based on searching a DT of MAP [11]. The second approach is based on the idea of clustering DTs in the two-dimensional space of DT size and DT fitness [12].

Our approach is based on the quantitative estimates of classification confidence, which can be made within the Uncertainty Envelope technique described in [9]. The idea behind our method of interpreting the Bayesian DT ensemble is to find a single DT which covers most of the training examples classified as confident and correct. For multiple classification systems the confidence of classification outputs can be easily estimated by counting the consistency of the classification outcomes.

Indeed, within a given classification scheme the outputs of the multiple classifier system depend on how well the classifiers were trained and how representative were the training data. For a given data sample, the consistency of classification outcomes depends on how close this sample is to the class boundaries. So for the *i*th class, the confidence in the set of classification models can be estimated as a ratio $\gamma_i$ between the number of classifier outcomes of the *i*th class, $N_i$, and the total number of classifiers $N$: $\gamma_i = N_i/N$, $i = 1, …, C$, where $C$ is the number of classes.

Clearly the classification confidence is maximal, equal to 1.0, if all the classifiers assign a given input to the same class, otherwise the confidence is less than 1.0. The minimal value of confidence is equal to $1/C$ if the classifiers assign the input datum to the $C$ classes in equal proportions. So for a given input the classification confidence in the set of classifiers can be properly estimated by the ratio $\gamma$.

Within the above framework in real-world applications, we can define a given level of the classification confidence, $\gamma_0$: $1/C \leq \gamma_0 \leq 1$, for which cost of misclassification is small enough to be accepted. Then for the given input, the outcome of the set of classifiers is said to be *confident* if the ratio $\gamma \geq \gamma_0$. Clearly, on the labeled data we can distinguish between *confident and correct* outcomes and *confident but incorrect* outcomes. The latter outcomes may appear in a multiple classifier system due to noise or overlapping classes in the data.

*3.2. A Selection Procedure*

In practice, the number of DTs in the ensemble as well as the number of the training examples can be large. Nevertheless, counting the number of confident and correct outcomes as described above, we can find a desired DT which can be used for interpreting the confident classification. The performance of such a DT can be slightly worse than that of the Bayesian DT ensemble. Within the Chapter we provide the experimental comparison of their performances. The main steps of the selection procedure are next.

All that we need is to find a set of DTs which cover the maximal number of the training samples classified as confident and correct while the number of misclassifications on the remaining examples is kept minimal. To find such a DT set, we can remove the conflicting examples from the training data and then select the DTs with a maximal cover of the training samples classified by the DT ensemble as confident and correct.

Thus the main steps of the selection procedure are as follows:

1. Amongst a given Bayesian DT ensemble find a set of DTs, S1, which cover a maximal number of the training samples classified as confident and correct with a given confidence level $\gamma_0$.
2. Find the training samples which were misclassified by the Bayesian DT ensemble and then remove them from the training data. Denote the remaining training samples as D1.
3. Amongst the set S1 of DTs find those which provide a minimal misclassification rate on the data D1. Denote the found set of such DTs as S2.
4. Amongst the set S2 of DTs select those whose size is minimal. Denote a set of such DTs as S3. The set S3 contains the desired DTs.

The above procedure finds one or more DTs and puts them in the set S3. These DTs cover a maximal number of the training samples classified as confident and correct with a given confident level $\gamma_0$. The size of these DTs is minimal and any of them can be finally selected for interpreting the confident classification.

**4. Experimental Results**

In this Section first we describe the data used in our experiments. Then we show how the suggested Bayesian technique runs on these data. The resultant Bayesian averaging over DT models gives us a feature importance diagram. The suggested selection procedure gives us the single DTs for each run and finally we compare the predictive accuracies obtained with the existing procedures.

*4.1. The Experimental Data*

The data used in our experiments are related to the Short-Term Conflict Alert (STCA) problem which emerges when the distance between two aircraft, landing or taking off, might be critically short. Table 1 lists 12 features selected for predicting STCA.

In this table $\Delta_x$, $\Delta_y$, and $\Delta_z$ are the distances between pairs of aircraft on the X, Y, and height Z axes, respectively. Feature $x_4 = \sqrt{\Delta_x^2 + \Delta_y^2 + \Delta_z^2}$ is the distance between pairs of aircraft in 3-dimensional space. $V_{x,1}$ is the velocity of craft 1 on axis X, …, $V_{z,2}$ is the velocity of craft 2 on height Z. $T_1$ and $T_2$ are the times since the last correlated plot in the lateral plane for aircraft 1 and aircraft 2, respectively.

In our experiments we used 2500 examples of radar cycles taken each 6 seconds. From these examples 984 cycles were labeled as alerts. The number of cycles relating to one pair is dependent on the velocity and, on average, is around 40. All the examples of radar cycles were split into halves for training and test data sets within 5 fold cross-validation.

**Table 1.** The features selected for predicting the STCA.

| # | Name | Feature |
|---|---|---|
| 1 | $x_1$ | $\Delta_x$ |
| 2 | $x_2$ | $\Delta_y$ |
| 3 | $x_3$ | $\Delta_z$ |
| 4 | $x_4$ | $\sqrt{\Delta_x^2 + \Delta_y^2 + \Delta_z^2}$ |
| 5 | $x_5$ | $V_{x,1}$ |
| 6 | $x_6$ | $V_{y,1}$ |
| 7 | $x_7$ | $V_{z,1}$ |
| 8 | $x_8$ | $V_{x,2}$ |
| 9 | $x_9$ | $V_{y,2}$ |
| 10 | $x_{10}$ | $V_{z,2}$ |
| 11 | $x_{11}$ | $T_1$ |
| 12 | $x_{12}$ | $T_2$ |

Fig. 1 shows two pairs of aircraft flying with different velocities: the distance between the aircraft is shown here by $x_4$ versus the radar cycles. The alert cycles are shown as stars, and the cycles, recognized by experts as normal, are shown as circles.

From Fig. 1, we can see that the left hand trace seems more complex for predicting the STCA than the right hand trace. First the series of alert cycles on the left hand trace is disrupted by 2 normal cycles, and second the aircraft having passed a critical 30[th] cycle remain in the alert zone. In contrast, the right hand trace seems straight and predictable.

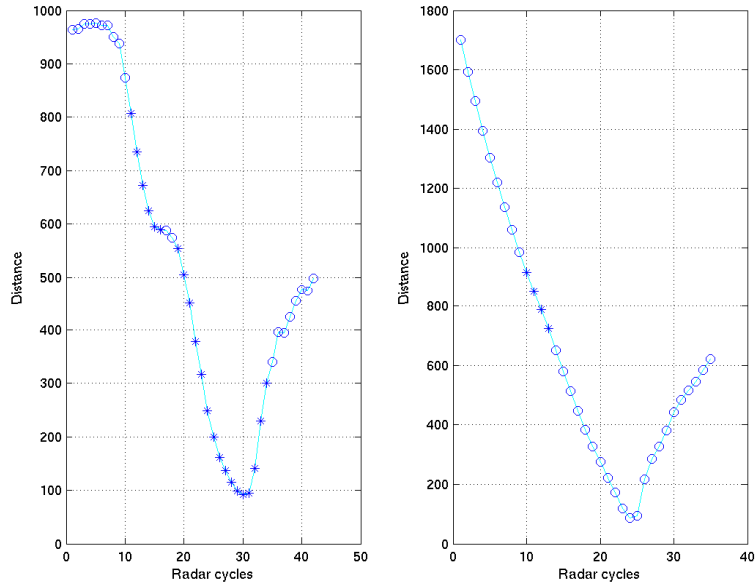

**Figure 1.** Two examples of alert cycles denoted here by the stars.

*4.2. Performance of the Bayesian DT averaging technique*

We ran the Bayesian DT technique without *a priori* information on the preferable DT shape and size. The minimal number of data points allowed in the splits, $p_{min}$, was set equal to 15 or 1.2% of the 1250 training examples. The proposal probabilities for the death, birth, change-split and change-rules were set to 0.1, 0.1, 0.2, and 0.6, respectively. The numbers of burn-in and post burn-in samples were set equal to 100k and 10k, respectively. The sampling rate was set equal to 7, and the proposal variance was set at 0.3 in order to achieve the rational rate of acceptance rate around 0.25, which was recommended in [5].

5 fold cross-validation was used to estimate the variability of the resultant DTs. The performances of all the 5 runs were nearly the same, and for the first run Fig. 2 depicts samples of log likelihood and numbers of DT nodes as well as the densities of DT nodes for burn-in and post burn-in phases.

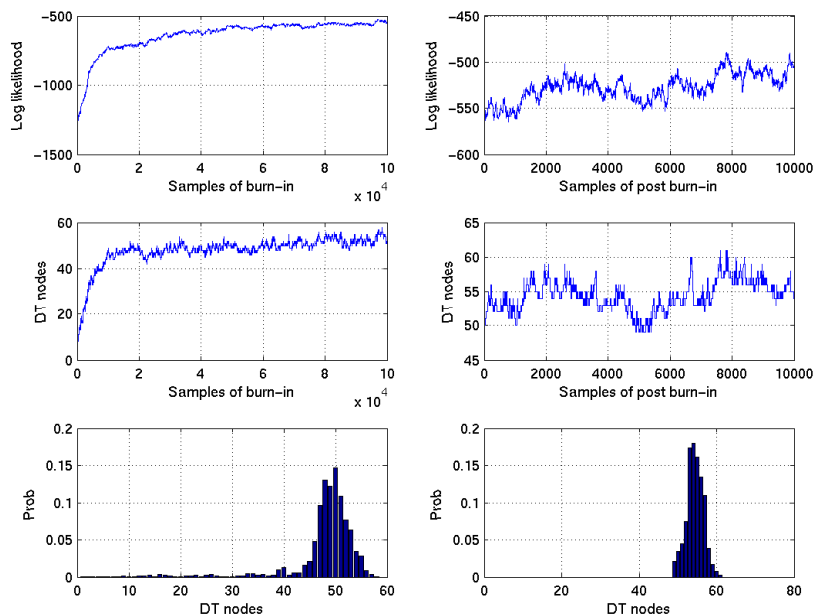

**Figure 2.** Samples of log likelihood and DT size during burn-in (the left side) and post burn-in (the right side). The bottom plots are the distributions of DT sizes.

From the top left plot we can see that the Markov chain converges to the stationary value of log likelihood near to –500 after starting around –1200. During the post burn-in phase the values of log likelihood slightly oscillate between –550 and –500.

The acceptance rates were 0.24 for the burn-in and 0.22 for the post burn-in phases. The average number of DT nodes and its variance were equal to 54.4 and 2.2, respectively.

On the first run, the Bayesian DT averaging technique misclassified 14.3% of the test examples. The rate of the confident and correct outcomes was 62.77%.

*4.3. Feature Importance*

Table 2 lists the average posterior weights of all the 12 features sorted by value. The bigger the posterior weight of a feature, the greater is its contribution to the outcome. On this basis, Table 2 provides ranks for all the 12 features.

Fig. 3 shows us the error bars calculated for the contributions of the 12 features to the outcome averaged over the 5 fold cross-validation. From this figure we can see that such features as $x_8$, $x_1$, and $x_9$ are used in the Bayesian DTs, on average, more frequently than the others. In contrast, feature $x_{12}$ is used with a less frequency. Additionally, the widths of the error bars in Fig. 3 give us the estimates of variance of the contributions.

**Table 2.** Posterior weights of the features sorted on their contribution to the outcome.

| Feature | Posterior weight | Rank |
|---|---|---|
| $x_8$ | 0.168 | 1 |
| $x_1$ | 0.137 | 2 |
| $x_9$ | 0.120 | 3 |
| $x_6$ | 0.110 | 4 |
| $x_4$ | 0.095 | 5 |
| $x_5$ | 0.090 | 6 |
| $x_3$ | 0.078 | 7 |
| $x_2$ | 0.061 | 8 |
| $x_{10}$ | 0.050 | 9 |
| $x_7$ | 0.042 | 10 |
| $x_{11}$ | 0.001 | 11 |
| $x_{12}$ | 0.008 | 12 |

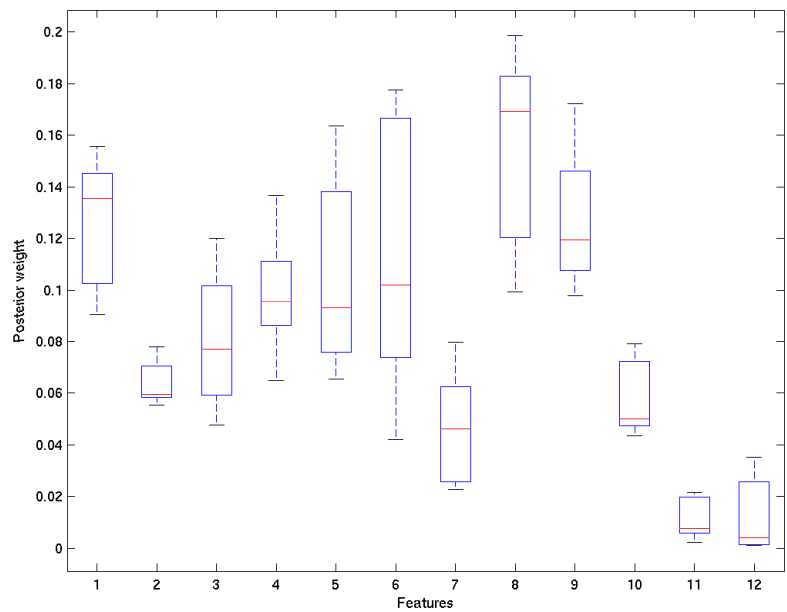

**Figure 3.** Feature importance averaged over 5 fold cross-validation.

*4.4. A Resultant DT*

The resultant DT selected by the SC procedure is presented as a machine diagram in Fig. 4. Each splitting node of the DT provides a specific question that has a yes/no answer, and two branches. The terminal nodes provide the predictive probabilities of alert, whose values range between 0.0 and 1.0.

```
node01   X04 < 1847.05, then node03, otherwise node45
node03   X04 < 1459.91, then node06, otherwise node28
node06   X05 < -281.95, then node15, otherwise node07
node07   X03 < 1713.61, then node08, otherwise node12
```

```
node08   X01 <   -1.64, then node02, otherwise node04
node02   X07 <    6.19, then node10, otherwise node14
node04   X04 <  324.47, then node18, otherwise node11
node10   X08 <   69.80, then node20, otherwise alert(0.99)
node11   X10 <  -13.43, then node19, otherwise node17
node12   X08 < -105.10, then alert(1.00), otherwise node09
node14   X08 <  150.30, then node23, otherwise alert(0.45)
node17   X05 <   82.84, then node25, otherwise node13
node18   X06 < -235.08, then alert(0.09), otherwise node31
node19   X08 <  -45.98, then node43, otherwise node05
node20   X04 <  415.29, then alert(0.13), otherwise node21
node21   X09 <   31.87, then alert(0.89), otherwise node39
node15   X09 <   81.89, then node34, otherwise node29
node25   X09 < -138.94, then node27, otherwise node41
node13   X01 <    2.31, then node44, otherwise node22
node22   X06 < -275.55, then alert(0.50), otherwise alert(0.99)
node28   X08 <  -28.49, then node16, otherwise node30
node29   X08 <  -46.49, then alert(0.06), otherwise alert(1.00)
node27   X05 <   11.31, then node42, otherwise alert(0.00)
node34   X01 <   -1.24, then alert(0.96), otherwise node32
node05   X11 <    0.00, then alert(0.07), otherwise node33
node31   X09 < -317.08, then alert(0.68), otherwise node37
node30   X03 < 4075.28, then alert(0.86), otherwise alert(0.00)
node39   X05 <  142.61, then alert(0.98), otherwise alert(0.27)
node37   X05 < -212.06, then node26, otherwise node40
node42   X02 <   -1.28, then node38, otherwise node51
node43   X01 <   -0.02, then alert(0.38), otherwise alert(1.00)
node16   X07 <   29.42, then node24, otherwise alert(0.25)
node41   X08 <  314.03, then alert(0.97), otherwise alert(0.56)
node33   X08 <   76.40, then alert(0.16), otherwise alert(0.86)
node40   X09 <  174.57, then node35, otherwise alert(0.12)
node38   X12 <    0.00, then alert(0.08), otherwise alert(1.00)
node35   X02 <    0.28, then alert(0.27), otherwise alert(0.79)
node51   X06 <   23.34, then alert(0.65), otherwise alert(1.00)
node44   X05 <  216.38, then alert(1.00), otherwise alert(0.62)
node45   X01 <  -15.93, then alert(0.96), otherwise alert(0.89)
node32   X01 <   13.80, then alert(0.00), otherwise alert(0.61)
node09   X10 <    0.64, then alert(0.89), otherwise node49
node23   X09 <   26.46, then alert(0.17), otherwise alert(0.07)
node49   X01 <    9.91, then alert(0.24), otherwise alert(0.57)
node24   X09 <  -99.68, then alert(1.00), otherwise alert(0.83)
node26   X05 < -239.50, then alert(0.06), otherwise alert(0.00)
```

**Figure 4.** Machine diagram of the resultant DT selected by the SC technique.

*4.5. Comparison of Performances*

In this section we compare our technique of extracting a sure correct (SC) DT with MAP, and the maximum a posterior weight (MAPW). The comparison is made in terms of misclassification within 5 fold cross-validation. The misclassification rates of the above three techniques: SC, MAP, and MAPW are shown in Fig. 5. The left side plot shows the misclassification rates of the single DTs on the test data, and the right side plot shows its sizes.

In theory, the Bayesian averaging technique should provide lower misclassification rates than any other single DTs selected by the SC, MAP, and MAPW techniques. On the first run, we can observe that all the single DTs perform worse than the Bayesian ensemble of DTs which has misclassified 14.3% of the test data.

Comparing the misclassification rates of the SC, MAP, and MAPW shown in Fig. 5, we can see that the suggested SC technique more often out-performs the other two techniques, that is, the SC technique out-performs the MAP and MAPW techniques on the 4 runs.

Comparing the DT sizes on the right side plot, we can see that the SC technique has extracted shorter DTs than the MAP technique in 4 runs. At the same time, comparing the sizes of the SC and MAPW DTs, we can see that the SC technique has extracted shorter DTs in 2 runs only.

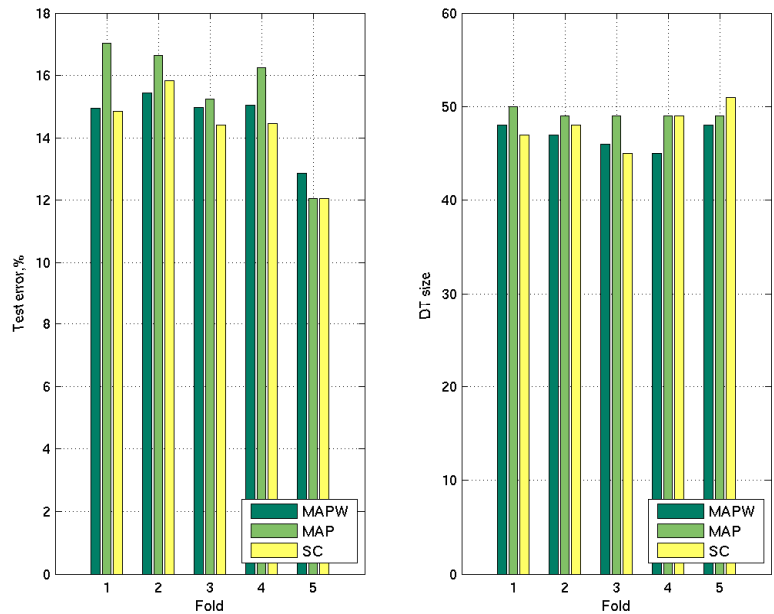

**Figure 5.** Comparison of test error and DT sizes within 5 fold cross-validation for the MAPW, MAP and proposed SC techniques.

## 5. An Application Scenario

In a general form an application scenario within our approach can be described as a sequence of the following steps.

1. Define a set of features considered by domain experts as the most important ones for the classification. For example, a domain expert can assume the velocities of aircraft in coordinates X and Y as the most important features for predicting the STCA.
2. Within a defined set of features, collect a representative set of data samples confidently classified by domain experts. For instance, a domain expert can arrange a set of radar data received from different pairs of aircraft in which some of the radar cycles were labeled as the STCA.
3. Analyze *a priori* knowledge and formulate priors within the Bayesian methodology. For example, domain experts can represent their knowledge in a form of decision tree asking specific questions. Such *a priori* information can be used within our approach in order to improve the performance.
4. Define parameters of DTs such as the minimal number of data samples, $p_{min}$, allowed to be in splitting nodes. By changing this parameter, a modeler can find in an *ad hoc* manner the number of splitting nodes providing the best performance of DTs.
5. Define parameters of MCMC such as the number of burn-in and post burn-in samples, proposal probabilities for the death, birth, change-split and change-rule, as well as the sampling rate. At this step a modeler can also specify suitable proposal distributions for the moves. However, if a modeler has no idea about the proposal distribution, a uniform distribution, known also as an "uninformative" prior, is used.
6. Specify a criterion for convergence of Markov Chain. Within the Bayesian MCMC technique, the convergence of a Markov Chain is usually achieved if the number of burn-in samples set enough large. Practically, the convergence is achieved if after approximately 1/3 of burn-in samples the likelihood values do not change significantly. This can be easily visualized by observing the log likelihood values plotted versus the number of post burn-in samples.
7. A modeler has to control the diversity of DTs collected during the post burn-in phase. The diversity is estimated implicitly by estimating an acceptance level which is the ratio of the number of the proposed and accepted DTs to the total number of proposed DTs. Practically, when the acceptance level is near 0.25, a set of collected DTs is optimally diversified.
8. Practically, the desired acceptance level is achieved by changing the following parameters: $p_{min}$, the variance of proposal distribution, and the expected number of splitting nodes.
9. To select a single DT providing the most confident classification, a modeler has to predefine a level of confident classification, $\gamma_0$. The value of $\gamma_0$ is dependent on the cost of misclassifications allowed in an application. Clearly, if the cost of classification is high, the value of $\gamma_0$ is defined close to 1.0, say 0.9990. This means that a decision is confident if no more than 10 classifiers from 10000 are contradictory. Otherwise, a decision is assigned uncertain.
10. Having obtained a confident DT, a modeler can observe a decision model and analyze the features used in this model. Additionally, a modeler can run the Bayesian MCMC and DT selection techniques within *n*-fold cross validation in order to estimate the contribution of each feature to the classification. As a result, the contributions of all features can be ranked, and a modeler can find the most critical features in an application.

Related to the STCA problem, the application scenario used in our work is described as follows. The domain experts have defined 12 features listed in Table 1. From the collected 2500 data samples, 1250 were used for training and the remaining 1250 for testing. *A priori* information was not available in our case and, therefore, priors were given as "uninformative" except for a Gaussian for the proposal distribution with the variance set to 0.3. The number $p_{min}$ = 15 was experimentally found to provide the best performance. The numbers of burn-in and post burn-in samples were set equal to 100 k and 10 k, respectively. The proposal probabilities for the death, birth, change-split and change-rules were set to be 0.1, 0.1, 0.2, and 0.6, respectively. Every $7^{th}$ DT was collected during the post burn-in phase, i.e., the sampling rate was 7. The convergence of the Markov Chain can be visually observed from the top right plot in Fig. 2 – from this plot we can see that after approximately 1/3 of burn-in samples the values of log likelihood do not change significantly. The acceptance level during the post burn-in phase was obtained equal to 0.22 that is close to 0.25 when the diversity of DTs is optimal. The level of confident classification, $\gamma_0$, was predefined to be 0.99. Fig 4 represents the machine diagram of the resultant DT selected under the given $\gamma_0$. Finally, the performance of the resultant DT is compared with the performances of the Bayesian DT technique as well as the MAP DT within the 5 fold cross-validation as shown in Fig. 5.

## 6. Conclusion

For estimating uncertainty of decisions in safety-critical engineering applications, we have suggested the Bayesian averaging over decision models using a new strategy of the RJ MCMC sampling for the cases when *a priori* information on the favored structure of models is unavailable. The use of DT models assists experts to interpret causal relations and find factors to account for the uncertainty. However, the Bayesian averaging over DTs allows experts to estimate the uncertainty accurately when *a priori* information on favored structure of DTs is available.

To interpret an ensemble of diverse DTs sampled by the RJ MCMC technique, experts select the single DT model that has maximum *a posteriori* probability. However in practice this selection technique tends to choose over-fitted DTs which are incapable of providing a high predictive accuracy.

In this Chapter we have proposed a new procedure of selecting a single DT. This procedure is based on the estimates of uncertainty in the ensemble of the Bayesian DTs. For estimating the uncertainty, the use of an Uncertainty Envelope technique has been advocated. As a result, in our experiments with the STCA data, the suggested technique outperforms the existing Bayesian techniques in terms of predictive accuracy.

Thus, we conclude that the technique proposed for interpreting the ensemble of DTs allows experts to select a single DT providing the most confident estimates of outcomes. These are very desirable properties for classifiers used in safety-critical systems, in which assessment of uncertainty of decisions is of crucial importance.

## 7. Acknowledgements

The work reported was largely supported by a grant from the EPSRC under the Critical Systems Program, grant GR/R24357/01.

**References**

[1]  A. Haldar, S. Mahadevan, *Probability and Statistical Methods in Engineering Design*, Wiley, 1999.
[2]  R. Everson, J.E. Fieldsend, Multi-Objective optimization of safety related Systems: An application to short term conflict alert, *IEEE Transactions on Evolutionary Computation* (forthcoming) 2006.
[3]  L. Kuncheva, *Combining Pattern Classifiers: Methods and Algorithms*, Wiley, 2004.
[4]  L. Brieman, J. Friedman, R. Olshen, C. Stone, *Classification and Regression Trees*, Belmont, CA, Wadsworth, 1984.
[5]  D. Denison, C. Holmes, B. Malick, A. Smith, *Bayesian Methods for Nonlinear Classification and Regression*, Wiley, 2002.
[6]  H. Chipman, E. George, R. McCullock, Bayesian CART model search, *J. American Statistics* **93** (1998), 935-960.
[7]  P. Domingos, Bayesian averaging of classifiers and the overfitting problem, *International Conference on Machine Learning. Stanford, CA, Morgan Kaufmann* 2000, 223-230.
[8]  V. Schetinin, J. E. Fieldsend, D. Partridge, W. J. Krzanowski, R. M. Everson, T. C. Bailey, and A. Hernandez, The Bayesian decision tree technique with a sweeping strategy, *Int. Conference on Advances in Intelligent Systems - Theory and Applications, (AISTA'2004) in cooperation with IEEE Computer Society,* Luxembourg, 2004.
[9]  J.E. Fieldsend, T.C. Bailey, R.M. Everson, W.J. Krzanowski, D. Partridge, V. Schetinin, Bayesian inductively learned modules for safety critical systems, *Symposium on the Interface: Computing Science and Statistics*, Salt Lake City, 2003.
[10]  P. Green, Reversible jump Markov Chain Monte Carlo computation and Bayesian model determination, *Biometrika* **82** (1995), 711-732.
[11]  P. Domingos, Knowledge discovery via multiple models. *Intelligent Data Analysis* **2** (1998), 187-202.
[12]  H. Chipman, E. George, R. McCulloch, Making sense of a forest of trees, *Symposium on the Interface*, S. Weisberg, Ed., Interface Foundation of North America, 1998.